\documentclass[a4paper]{article}
\usepackage{algorithm, algorithmic}
\usepackage{graphicx}
\usepackage{subcaption}
\usepackage{amsmath}
\usepackage{booktabs}
\usepackage{url}
\usepackage{authblk}

\title{Guiding Application Users via Estimation of Computational Resources for Massively Parallel Chemistry Computations}

\date{}

\author[1]{Tanzila Tabassum}
\author[2]{Omer Subasi}
\author[3]{Ajay Panyala}
\author[4]{Epiya Ebiapia}
\author[5]{Gerald Baumgartner}
\author[6]{Erdal Mutlu}
\author[7]{P. (Saday) Sadayappan}
\author[8]{Karol Kowalski}

\affil[1,4,5]{Louisiana State University, Louisiana, USA}
\affil[2,3,6,8]{Pacific Northwest National Laboratory, Washington, USA}
\affil[7]{University of Utah, Utah, USA}

\begin{document}

\maketitle

\begin{abstract}
In this work, we develop machine learning (ML) based strategies to predict resources (costs) required for massively parallel chemistry computations, such as coupled-cluster methods, to guide application users before they commit to running expensive experiments on a supercomputer. By predicting application execution time, we determine the optimal runtime parameter values such as number of nodes and tile sizes. Two key questions of interest to users are addressed. The first is the {\em shortest-time} question, where the user is interested in knowing the parameter configurations (number of nodes and tile sizes) to achieve the shortest execution time for a given problem size and a target supercomputer. The second is the {\em cheapest-run} question in which the user is interested in minimizing resource usage, i.e., finding the number of nodes and tile size that minimizes the number of node-hours for a given problem size.

We evaluate a rich family of ML models and strategies, developed based on the collections of  runtime parameter values for the CCSD (Coupled Cluster with Singles and Doubles) application executed on the Department of Energy (DOE) Frontier and Aurora supercomputers. 
Our experiments show that when predicting the total execution time of a CCSD iteration, a Gradient Boosting (GB) ML model achieves a Mean Absolute Percentage Error (MAPE) of 0.023 and 0.073 for Aurora and Frontier, respectively. In the case where it is expensive to run experiments just to collect data points, we show that active learning can achieve a MAPE of about 0.2 with just around 450 experiments collected from Aurora and Frontier.
\end{abstract}

\section{Introduction}
The development of highly scalable computational chemistry software  leads to successful research missions in computational and theoretical chemistry. Massively parallel computational chemistry applications are designed to leverage supercomputers to achieve high scalability and performance. A machine learning (ML) library offering accurate runtime predictions will be of significant benefit to the application users of a supercomputing system to optimize their use of the resources. By predicting the total runtime, one can answer the following user inquiries: i) Given a problem size and a target supercomputer for which there is sufficient historical data, how many nodes and what tile size should we use to minimize the execution time? ii) If the objective is to minimize resources used, what node count and tile size should be used? iii) What if a user does not have much historical data for the target application and supercomputer?  In that case, if a small amount of data could be collected, what problem sizes, the numbers of nodes, and the tile sizes should be selected to test-run the application such that the accuracy of the runtime prediction improves the most? 

Therefore, in this work, we develop various ML models and strategies for massively parallel computational chemistry calculations so as to guide the application users and respond to their inquiries about computational resources. In particular, we focus on the Coupled Cluster with Singles and Doubles (CCSD)~\cite{ccsd1,ccsd2,ccsd3} application and two types of inquiry of great interest. The Shortest-Time Question (STQ) asks how many nodes should be used and what should be the application parameter configuration values (such as tile size) to achieve the shortest possible execution time? 
The budget question (BQ) asks for the parameter configuration that minimizes the user's budget expenditure in terms of node-hours to execute their application. 

In addressing the modeling problem, we consider two real-world scenarios. The first is that the user has a sufficient amount of historical experimental data in the form of  execution time values for various problem sizes, numbers of nodes, and tile sizes. For this scenario, we train and evaluate a rich suite of ML models with cross validation. 
The second scenario is when the historical experimental data does not exist and it is expensive to run experiments just to collect data points. To address this scenario, we utilize active learning.
In our active learning-based approach, we explore two types of query strategies: uncertainty sampling and query by committee.

We find that, in general, Gradient Boosting regression performs the best in terms of mean absolute error (MAE), mean absolute percentage error (MAPE), and the coefficient of determination ($R^2$ score). For the experiments that were run on ALCF Aurora~\cite{alcf_aurora}, when predicting the total execution time, our Gradient Boosting model achieves an $R^2$, MAE, and MAPE score of 0.999, 2.36, and 0.02, respectively. On OLCF Frontier~\cite{olcf_frontier}, it achieves an $R^2$, MAE, and MAPE score of 0.969, 4.65, and 0.073 respectively.

The paper is organized as follows:
In Section \ref{background}, 
we provide background and discuss related work.
In Section \ref{sec:ourproposal}, we present our approach. 
In Section \ref{experimental_results}, we describe the experimental setup and
discuss the experimental results.
Finally, we conclude our work with Section \ref{conclusions}.

\section{Background and Related Work}
\label{background}
Modern computational chemistry applications, particularly those based on correlated electronic structure methods such as Coupled Cluster with Singles and Doubles (CCSD)~\cite{ccsd1,ccsd2,ccsd3}, are among the most demanding scientific workloads run on leadership-class supercomputers. The rapid increase in available computational power has enabled calculations on increasingly large and complex molecular systems, but it has also created significant challenges for efficiently allocating computational resources and choosing optimal runtime parameters. This is especially true in the context of massively parallel codes where runtime performance is a complex function of problem size, hardware configuration, and application-specific parameters.

Resource estimation for quantum chemistry methods has traditionally relied on theoretical scaling laws and empirical heuristics. While these methods provide general insights, they are insufficient for accurately predicting actual runtime or near optimal application parameters, e.g., block size, and hardware parameters, e.g., nodes and GPUs, on modern heterogeneous architectures. Factors such as communication overhead, load imbalance, and GPU acceleration introduce nontrivial effects that make performance estimation challenging.

ML techniques offer a promising alternative for performance modeling in high-performance computing (HPC). By learning from historical data, ML models can capture complex, nonlinear interactions between input features (e.g., number of occupied and virtual orbitals, number of nodes, tile sizes) and target metrics (e.g., runtime, memory usage, cost). Applications of ML in HPC have included runtime prediction, power modeling, fault detection, and auto-tuning of compiler flags or runtime parameters.

To address the challenge of limited training data on emerging architectures, active learning is explored in this work. Active learning selects the most informative data points to label, reducing the number of required samples. Approaches such as uncertainty sampling~\cite{settles2009active,tran2022active} and query-by-committee~\cite{seung1992query} have shown promise in selecting optimal configurations to train predictive models efficiently.

In this work, we build upon and integrate these ideas to create a robust, ML-driven framework that can answer key user-facing questions about performance and cost before a simulation is run. Our approach combines a diverse suite of regression models, including both classical and deep learning techniques, with active learning strategies to adapt to different data availability scenarios. We validate our methodology using data collected from CCSD runs on two leadership class supercomputing platforms: ALCF Aurora~\cite{alcf_aurora} and OLCF Frontier~\cite{olcf_frontier}.

\section{A Framework for Application Users}
\label{sec:ourproposal}
In this section, we present our ML models, the baseline approach and the active learning approach.

\subsection{Evaluated ML Models}
\label{sub:ml_models}
\textbf{Polynomial Regression (PR)} introduces polynomial terms to model nonlinear relationships. It is still linear in coefficients but nonlinear in features.

\textbf{Kernel ridge regression (KR)} combines ridge regression (linear least squares with l2-norm regularization) with the kernel trick. It thus learns a linear function in the space induced by the respective kernel and the data.

\textbf{Decision Trees (DTs)} split the feature space into rectangular regions by applying rules on individual features. A regression tree predicts the mean value of the target variable in each region (leaf node).
DTs are easy to interpret and can capture complex interactions among the features without requiring data normalization or standardization.
However, they are highly sensitive to data changes. 

\textbf{Random Forests (RFs)} are an ensemble method that builds multiple DTs using bootstrapping \cite{breiman2001random}.
RFs typically improve generalization by averaging the predictions of multiple DTs.
They require more computational resources and memory due to the storage and evaluation of multiple trees.

\textbf{Gradient Boosted Trees (GBs)} construct trees sequentially and each new tree starts learning by correcting the errors made by the previous ensemble. While GBs perform very well in general, 
they are sensitive to hyper-parameter choices. Experiments show that GBs are the best performing models for Aurora and Frontier. Thus, we employ them in our active learning approach.

\textbf{AdaBoosted Trees (ABs)} combine a sequential series of base learners  e.g., DTs, and starts training with the previously inaccurately predicted data points --- just as GBs. Different from GBs, ABs adjust the weights of the data points and the learners throughout their training. \cite{freund1997decision}. ABs are sensitive to noisy data and outliers.

\textbf{Gaussian Processes (GPs)} model distributions over functions using a kernel function. Predictions come with uncertainty estimates. GPs are fully Bayesian and provide posterior means and variances for predictions \cite{rasmussen2006gaussian}. GPs provide predictions with uncertainty estimates. However, 
they scale poorly with the number of training data points due to their cubic computational complexity. 

\textbf{Bayesian Ridge Regression (BR)} extends ridge regression with Bayesian inference. It assumes priors on model coefficients and computes posterior distributions. BR automatically estimates regularization parameters from data using marginal likelihood \cite{bishop2006pattern}. 

\textbf{Support Vector Regression (SVR)} is an extension of Support Vector Machines (SVM) that can be used to solve regression problems. It optimizes a function by finding a tube that approximates a continuous-valued function while minimizing the prediction error.

\subsection{Evaluation Metrics}
\label{sub:metrics}
$R^2$ score or Coefficient of Determination is a statistical measure that denotes how the proportion of the variance is explained by the independent variable.  This metric ranges from 0 to~1, where higher values indicate a better fit by the model:
$$R^2 = 1 - \frac{\sum_{i=1}^{n} (y_i - \hat{y}_i)^2}{\sum_{i=1}^{n} (y_i - \bar{y})^2}$$
where $y_i$ are the observed values, $\hat{y}_i$ are the predicted values, and $\bar{y}$ is the mean of the observed data.

Mean Absolute Error (MAE) is a scale-dependent metric that measures the average magnitude of absolute prediction errors, indicating how close predictions are to actual observations, regardless of direction. It is calculated as:
\begin{equation}\nonumber
\text{MAE} = \frac{1}{n}\sum_{i=1}^{n}\left| y_i - \hat{y}_i \right|
\end{equation}
where \( n \) is the total number of observations, \( y_i \) represents the observed values, and \( \hat{y}_i \) represents the predicted values.

Mean Absolute Percentage Error (MAPE) is a scale-independent metric that evaluates error as a percentage of the corresponding observed value. In cases where the scale or magnitude of the data points varies significantly, emphasizing the relative size of the error provides a better insight into the model performance.
 It is defined as:
\begin{equation}\nonumber
\text{MAPE} = \frac{1}{n}\sum_{i=1}^{n}\left|\frac{y_i - \hat{y}_i}{y_i}\right| \times 100\%
\end{equation}
Lower values of both MAE and MAPE indicate better performance.

\subsection{Supervised Learning}
\label{sub:baseline}
Our framework for guiding application users and answering 
their questions about the computational resources focuses on two distinct
real-world scenarios. In the first scenario, there are no restrictions on obtaining data points for training the predictive model. That is, we assume that we are able to run a sufficient number of tests (CCSD experiments in this work) on a target supercomputer and collect runtime parameter values (problem sizes, number of nodes and tile size) together with the total execution time to train a performant ML model. For this scenario, our baseline approach is mainly built on the exploration and evaluation of traditional (non-deep learning) ML models. We do not use deep learning as the traditional algorithms listed above make accurate predictions and are computationally cheaper than deep learning models. 

Our main goal is to answer the following two main questions: 
\textbf{Shortest-Time Question (STQ):} This is to inquire about the runtime parameter configurations, i.e., the set of parameter values, that will lead to
solving a target problem in the shortest time on a target supercomputer. In our study, an example for such a question would be: what parameters (number of nodes and tile size) should be used for the fastest possible execution of CCSD for a given problem size of $(O, V)$ (occupied and virtual orbitals) on Aurora?

\textbf{Budget Question (BQ):} This is to inquire about the parameter configurations that minimize the computational budget, e.g., node-seconds on Aurora to execute CCSD for a given problem size (in terms of $(O, V)$).

\def\NumNodes{\mbox{\em NumNodes}}
\def\TileSize{\mbox{\em TileSize}}
\def\user{\mbox{\small\em user}}
\def\trial{\mbox{\small\em trial}}

The modeling approach to answer both the above questions is to first build an ML regression model that predicts execution time for a given combination of $\langle O, V, \NumNodes, \TileSize\rangle$, where $O$ and $V$ represent the number of occupied and virtual orbitals of the physical system modeled,  $\NumNodes$ is the number of supercomputer nodes used for the job, and $\TileSize$ is a key blocking parameter that determines how large tensors are partitioned into smaller slices for distributed CCSD execution. This model is built with a training set containing around 2,000 instances, covering a range of problem sizes, tile-sizes, and number of nodes of typical use with the application.

The trained regression model to predict execution time is then used to answer the user questions of interest by iteratively using the model.
For a given $\langle O_{\user},V_{\user}\rangle$ corresponding to the physical system being modeled, the trained ML regression model is queried to predict the execution time for various combinations of $\langle O_{\user},V_{\user},\NumNodes_{\trial},\TileSize_{\trial}\rangle$, where $\langle O_{\user},V_{\user}\rangle$ is fixed and $\langle \NumNodes_{\trial},\TileSize_{\trial}\rangle$ are swept over a range of typical interest.

\begin{algorithm}[!ht]
\caption{Uncertainty Sampling (US) with Gaussian Processes}
\label{algorithm:us}
\begin{algorithmic}
\STATE $n_{\text{initial}} \leftarrow 50$
\STATE $\text{indices} \leftarrow \text{Randomly select } n_{\text{initial}} \text{ indices from } X_{\text{train}}$
\STATE $X_{\text{labeled}} \leftarrow X_{\text{train}}[\text{indices}]$;
$y_{\text{labeled}} \leftarrow y_{\text{train}}[\text{indices}]$
\STATE $X_{\text{unlabeled}} \leftarrow X_{\text{train}} \setminus 
 X_{\text{labeled}}$;
$y_{\text{unlabeled}} \leftarrow y_{\text{train}} \setminus 
 y_{\text{labeled}}$
\STATE $n_{\text{queries}} \leftarrow 20$;
\STATE $\text{query\_size} \leftarrow 50$
 $\text{r2\_list} \leftarrow []$;
 $\text{mape\_list} \leftarrow []$
\STATE $\text{mae\_list} \leftarrow []$
\STATE $is\_STQ \leftarrow \textbf{True/False} $
\IF{$is\_STQ$}
\STATE $\text{r2\_list\_stq} \leftarrow []$;
 $\text{mape\_list\_stq} \leftarrow []$;
 $\text{mae\_list\_stq} \leftarrow []$
\STATE $\text{optimal\_test} = \text{get\_optimal\_values}(X_{\text{test}}, y_{\text{test}})$
\ENDIF
\STATE $\text{model} \leftarrow \text{GaussianProcessRegression(...)}$
\FOR{$i = 1 \to n_{\text{queries}}$}
    \STATE $\text{model.fit}(X_{\text{labeled}}, y_{\text{labeled}})$
    \STATE $y_{\text{pred}}, \text{std} \leftarrow \text{model.predict}(X_{\text{train}}, \text{return\_std=True})$
    \STATE $\text{r2} \leftarrow \text{r2\_score}(y_{\text{train}}, y_{\text{pred}})$
    \STATE $\text{mape} \leftarrow \text{mean\_absolute\_percentage\_error}(y_{\text{train}}, y_{\text{pred}})$
    \STATE $\text{mae} \leftarrow \text{mean\_absolute\_error}(y_{\text{train}}, y_{\text{pred}})$
    \STATE $\text{r2\_list.append(r2)}$, $\text{mape\_list.append(mape)}$, $\text{mae\_list(mae)}$
    \STATE $ \_, \text{std}\leftarrow \text{model.predict}(X_{\text{unlabeled}},\text{return\_std=True})$
    \STATE $\text{q\_indices} \leftarrow \text{argsort}(-\text{std})[:\text{query\_size}]$
    \STATE $X_{\text{labeled}} \leftarrow X_{\text{labeled}} \cup X_{\text{unlabeled}}[\text{q\_indices}]$
    \STATE $y_{\text{labeled}} \leftarrow y_{\text{labeled}} \cup y_{\text{unlabeled}}[\text{q\_indices}]$
    \STATE $X_{\text{unlabeled}} \leftarrow X_{\text{unlabeled}} \setminus X_{\text{unlabeled}}[\text{q\_indices}]$
    \STATE $y_{\text{unlabeled}} \leftarrow y_{\text{unlabeled}} \setminus y_{\text{unlabeled}}[\text{q\_indices}]$
    \IF{$is\_STQ$}
    \STATE $y_{\text{pred}} \leftarrow \text{model.predict}(X_{\text{test}})$
    \STATE $\text{optimal\_pred} \leftarrow \text{get\_optimal\_values}(X_{\text{test}}, y_{\text{pred}})$
    \STATE $\text{r2, mae, mape} \leftarrow ...$ 
    \STATE $... \text{compute\_{losses} ($X_{\text{test}}$,  $y_{\text{test}}$  optimal\_test, optimal\_pred )}$
    \STATE $\text{r2\_list\_stq.append(r2)}$, $\text{mape\_list\_stq.append(mape)}$ 
    \STATE $\text{mae\_list\_stq(mae)}$
    \ENDIF
\ENDFOR
\end{algorithmic}
\end{algorithm}
\begin{algorithm}[!ht]
\caption{Query-by-Committee (QC) with Gradient Boosting}
\label{algorithm:qc}
\begin{algorithmic}
\STATE $n_{\text{initial}} \leftarrow 50$
\STATE $\text{indices} \leftarrow \text{Randomly select } n_{\text{initial}} \text{ indices from } X_{\text{train}}$
\STATE $X_{\text{labeled}} \leftarrow X_{\text{train}}[\text{indices}]$;
 $y_{\text{labeled}} \leftarrow y_{\text{train}}[\text{indices}]$
\STATE $X_{\text{unlabeled}} \leftarrow X_{\text{train}} \setminus X_{\text{labeled}}$;
 $y_{\text{unlabeled}} \leftarrow y_{\text{train}} \setminus y_{\text{labeled}}$
\STATE $n_{\text{committees}} \leftarrow 5$; 
 $n_{\text{queries}} \leftarrow 10$
\STATE $\text{query\_size} \leftarrow 50$
\STATE $\text{r2\_list} \leftarrow []$;
 $\text{mape\_list} \leftarrow []$;
 $\text{mae\_list} \leftarrow []$
\STATE $is\_STQ \leftarrow \textbf{True/False} $
\IF{$is\_STQ$}
\STATE $\text{r2\_list\_stq} \leftarrow []$;
 $\text{mape\_list\_stq} \leftarrow []$;
 $\text{mae\_list\_stq} \leftarrow []$
\STATE $\text{optimal\_test} = \text{get\_optimal\_values}(X_{\text{test}}, y_{\text{test}})$
\ENDIF
\FOR{$i = 1 \to n_{\text{queries}}$}
    \STATE $\text{committee\_predictions} \leftarrow ....$
    \STATE $... \text{zeros}(X_{\text{unlabeled}}.\text{shape}[0], n_{\text{committees}})$
    \FOR{$j = 1 \to n_{\text{committees}}$}
        \STATE $\text{model} \leftarrow 
        \text{GradientBoostingRegressor}(...)$
        \STATE $\text{model.fit}(X_{\text{labeled}}, y_{\text{labeled}})$
        \STATE $\text{committee\_predictions}[:, j] \leftarrow \text{model.predict}(X_{\text{unlabeled}})$
    \ENDFOR
    \STATE $\text{prediction\_variance} \leftarrow \text{{var}}(\text{committee\_predictions})$
    \STATE $\text{q\_indices} \leftarrow \text{{argsort}}(-\text{prediction\_variance})[:\text{query\_size}]$
    \STATE $X_{\text{query}}, y_{\text{query}} \leftarrow X_{\text{unlabeled}}[\text{q\_indices}], y_{\text{unlabeled}}[\text{q\_indices}]$
    \STATE $X_{\text{labeled}} \leftarrow \text{vstack}([X_{\text{labeled}}, X_{\text{query}}])$
    \STATE $y_{\text{labeled}} \leftarrow \text{append}(y_{\text{labeled}}, y_{\text{query}})$
    \STATE $X_{\text{unlabeled}} \leftarrow X_{\text{unlabeled}} \setminus X_{\text{query}}$
    \STATE $y_{\text{unlabeled}} \leftarrow y_{\text{unlabeled}} \setminus y_{\text{query}}$
    \STATE $y_{\text{pred}} \leftarrow \text{model.predict}(X_{\text{train}})$
    \STATE $\text{r2} \leftarrow \text{r2\_score}(y_{\text{train}}, y_{\text{pred}})$
    \STATE $\text{mape} \leftarrow \text{mean\_absolute\_percentage\_error}(y_{\text{train}}, y_{\text{pred}})$
    \STATE $\text{mae} \leftarrow \text{mean\_absolute\_error}(y_{\text{train}}, y_{\text{pred}})$
    \STATE $\text{r2\_list.append(r2)}$, $\text{mape\_list.append(mape)}$, $\text{mae\_list(mae)}$
    \IF{$is\_STQ$}
    \STATE $y_{\text{pred}} \leftarrow \text{model.predict}(X_{\text{test}})$
    \STATE $\text{optimal\_pred} \leftarrow \text{get\_optimal\_values}(X_{\text{test}}, y_{\text{pred}})$
    \STATE $\text{r2, mae, mape} \leftarrow ...$
    \STATE $...\text{compute\_losses} (X_{\text{test}}, y_{\text{test}}, \text{optimal\_test, optimal\_pred})$
    \STATE $\text{r2\_list\_stq.append(r2)}$, $\text{mape\_list\_stq.append(mape)}$
    \STATE $\text{mae\_list\_stq(mae)}$
    \ENDIF
\ENDFOR
\end{algorithmic}
\end{algorithm}

\subsection{Active Learning}
\label{sub:active-learning}
The second real-world scenario is when there is very little historical data for an application on a supercomputer and it is very costly to run experiments on a target supercomputer to gather an adequate number of data points for training a model. To address this difficult scenario, we evaluate the effectiveness of active learning.

Active learning is a machine learning approach where the algorithm interactively queries an information source (or ``oracle'') to obtain labels for selected data instances. The goal is to identify the most informative points, thereby reducing the total number of queries and making the learning process more efficient. It is particularly useful in supervised settings where labeled data is scarce, costly, or time-consuming to obtain. In this work, we apply active learning to select the most informative parameter configurations for running CCSD\@. Since trial runs on a supercomputer can be expensive, minimizing the number of gathered training data points is desirable, while aiming to answer a specific scientific question (e.g., STQ or BQ).

There exist numerous query strategies in active learning. In this work, we select two strategies: In uncertainty sampling (US) (see Algorithm~\ref{algorithm:us}), one inquires about the data points 
for which the learning model is least certain about their outputs. Meanwhile, in the expected model change strategy, one inquires the data points causing the most change in the model (the change can be quantified by the gradient). In the expected error reduction change, one inquires about the data points which will likely reduce the model's generalization error the most. Finally, in query by committee (QC) (see Algorithm~\ref{algorithm:qc}), one inquires about the data points for which a committee of a variety of models trained on the currently available known (labeled) data disagree the most.  

For both algorithms, when the goal is STQ instead of just evaluating ML models, we determine the true and predicted optimal parameters from the true total execution time in the test set and the predicted total execution time. Care needs to be taken: We compute the  execution time losses based on the parameter configuration inferred from the predicted configuration not directly with the predicted execution time itself. While the losses computed with the predicted time itself will most often, if not always, be smaller than the losses inferred from the predicted configuration, it is \textbf{not} the true loss. 
We locate the predicted configuration and the corresponding \textbf{true} predicted time. This is critical for accurate evaluation of our models.

\section{Experimental Evaluation}
\label{experimental_results}
\subsection{Methodology and Experimental Setup}
\label{experimental_setup}
Our datasets were generated using a high-performance quantum chemistry application based on the Tensor Algebra for Many-body Methods (TAMM)~\cite{tamm_jcp_2023} framework. TAMM provides a distributed tensor algebra infrastructure designed for scalable and portable implementation of many-body methods such as Coupled Cluster with Single and Double excitations (CCSD)~\cite{ccsd1,ccsd2,ccsd3} on modern supercomputing platforms. We used the CCSD implementation in ExaChem~\cite{exachem_2023}, an open-source suite of scalable electronic structure methods. Exa\-Chem is built using the TAMM framework.

The computational problem for CCSD is defined by the number of molecular orbitals, which are typically divided into
occupied orbitals ($O$), which contain electrons in the reference wavefunction, and virtual orbitals ($V$), which are unoccupied orbitals that electrons can be excited into.
The total number of orbitals is denoted as $N = O + V$, and corresponds directly to the number of basis functions used to represent the electronic wavefunction of a molecule. The computational cost of CCSD scales approximately as $\mathcal{O}(O^2V^4)$, which means the time and memory required for a calculation increases rapidly with the number of electrons ($O$) and the number of virtual states ($V$). For this reason, $O$ and $V$ together define the size of the problem, and are often used as the primary features to describe input configurations when modeling or predicting performance.

 The rate-limiting step in CCSD calculations involves sextic-scaling tensor contractions, i.e., each CCSD iteration scales as $\mathcal{O}(O^2V^4)$. To model the performance of CCSD computations, we executed only a single iteration of the CCSD algorithm for each configuration
 since the performance characteristics (e.g., number of operations, runtime, memory usage, parallel efficiency) for each iteration is very stable and consistent for this application.

Each application run was configured with a unique set of problem sizes $(O, V)$ representing various molecular systems, and parallel execution parameters such as are number of nodes and tensor block (tile) sizes. TAMM efficiently schedules and executes the required tensor operations using its task-based runtime, which leverages MPI-based distributed execution and GPU acceleration through optimized vendor libraries.
The performance data collected from each run includes the metrics (runtime parameters), $O$, $V$, number of nodes, tile-size, and total wall time for a single iteration. 

\begin{table}[t]
    \centering
    \caption{Datasets and the corresponding size breakdowns.}
    \label{tab:datasets}
    \begin{tabular}{|l|c|c|c|} \hline
      System   & Total & Train & Test  \\ \hline
      Aurora     & 2329 &  1746 & 583 \\ \hline
      Frontier   & 2454  & 1840 & 614 \\ \hline
    \end{tabular}
\end{table}

All CCSD experiments were performed on DOE's leadership class systems ALCF Aurora~\cite{alcf_aurora} and OLCF Frontier~\cite{olcf_frontier}.
Table \ref{tab:datasets} shows the details about the size breakdown of our datasets into training and test sets. 

The Machine learning models were developed using Python (version 3.13.5), the pandas library (version 2.2.3), NumPy (version 2.1.3)~\cite{numpy} and the scikit-learn library (version 1.6.1)~\cite{sklearn}. We used the scikit-optimize library (version 0.10.2)~\cite{2020SciPy-NMeth} for hyperparameter optimization, and the figures were generated using Matplotlib (version 3.10.0)~\cite{matplotlib}. 

\begin{figure*}[ht]
    \centering
        \includegraphics[width=0.48\textwidth]{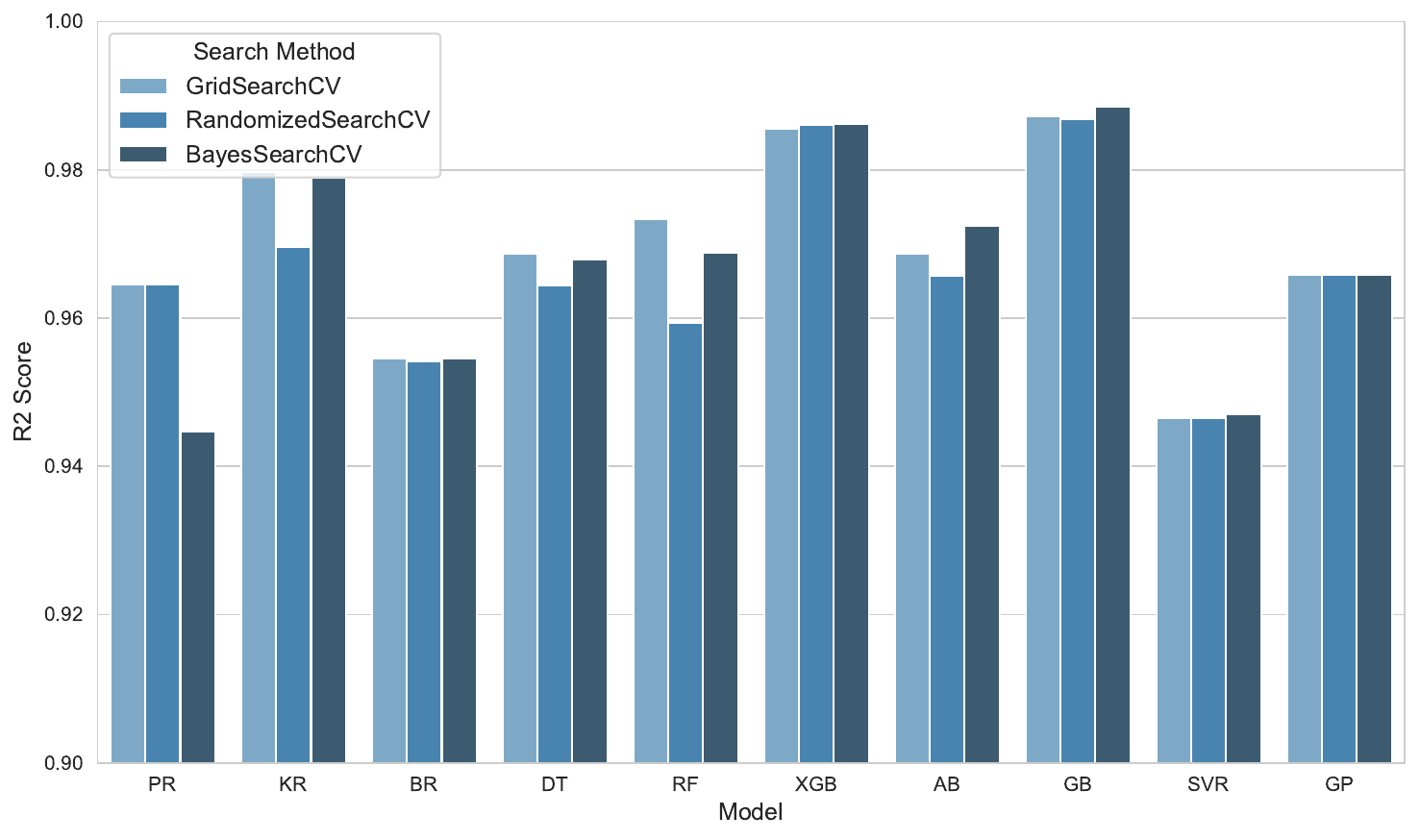}
        \includegraphics[width=0.48\textwidth]{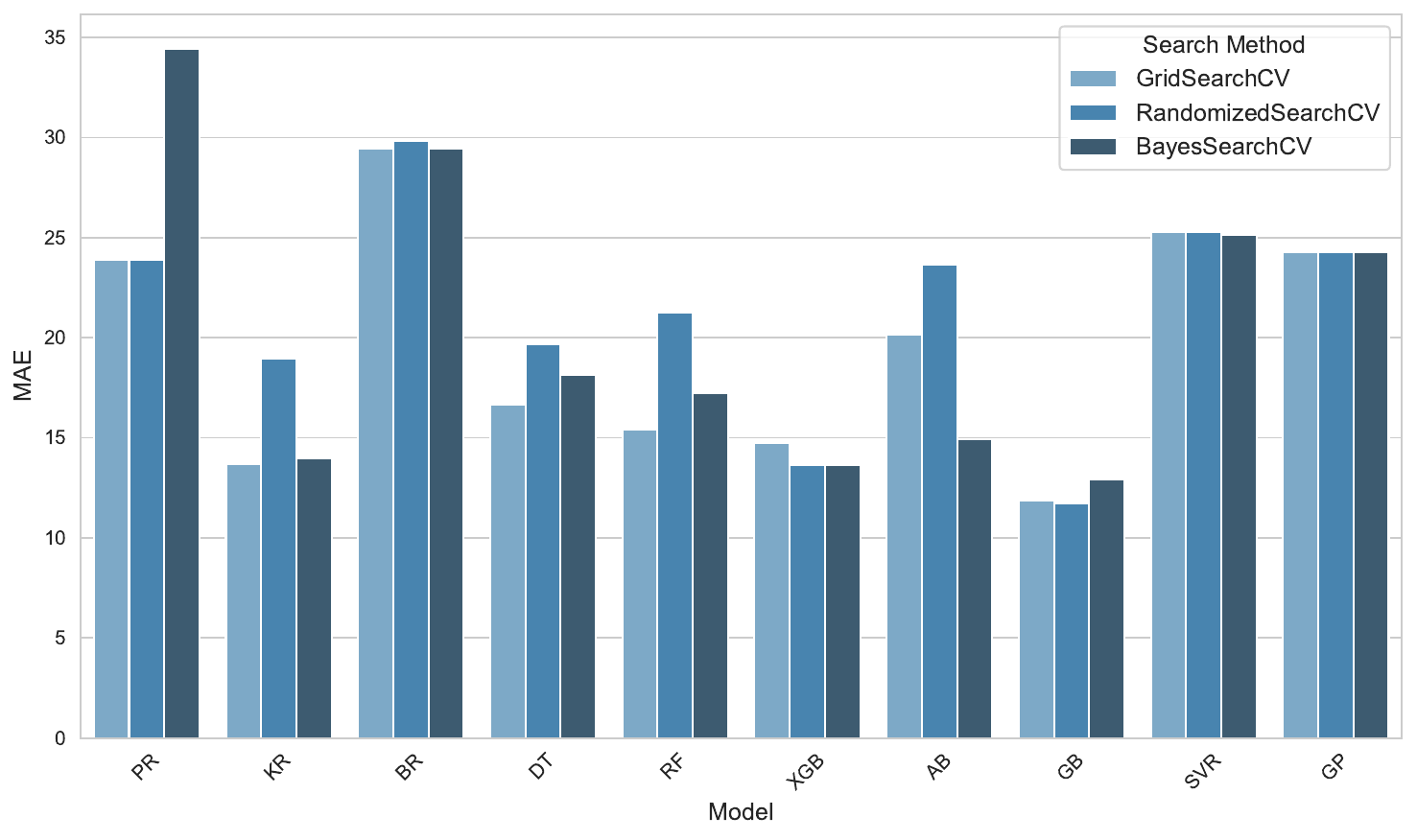}
        \includegraphics[width=0.48\textwidth]{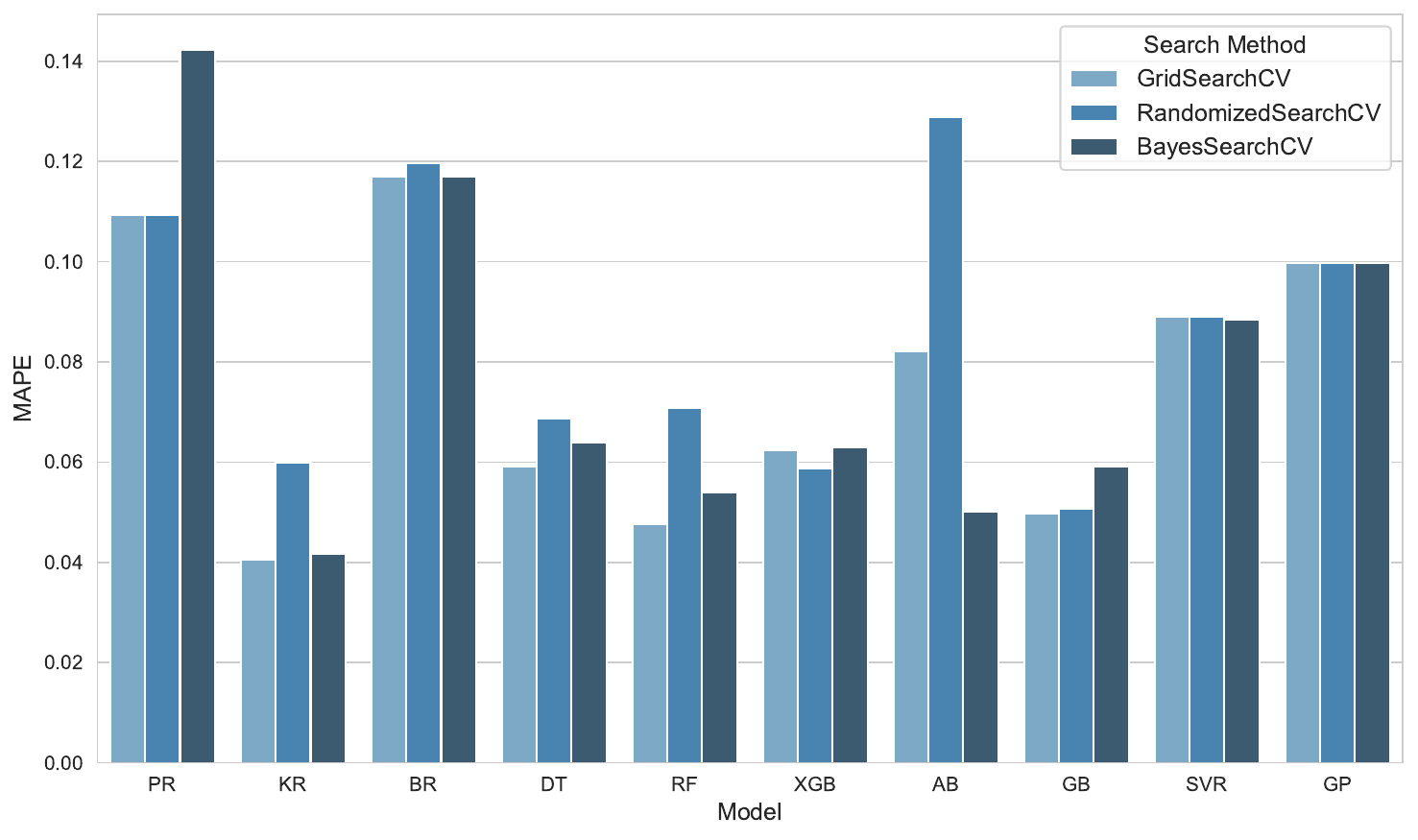}
        \includegraphics[width=0.48\textwidth]{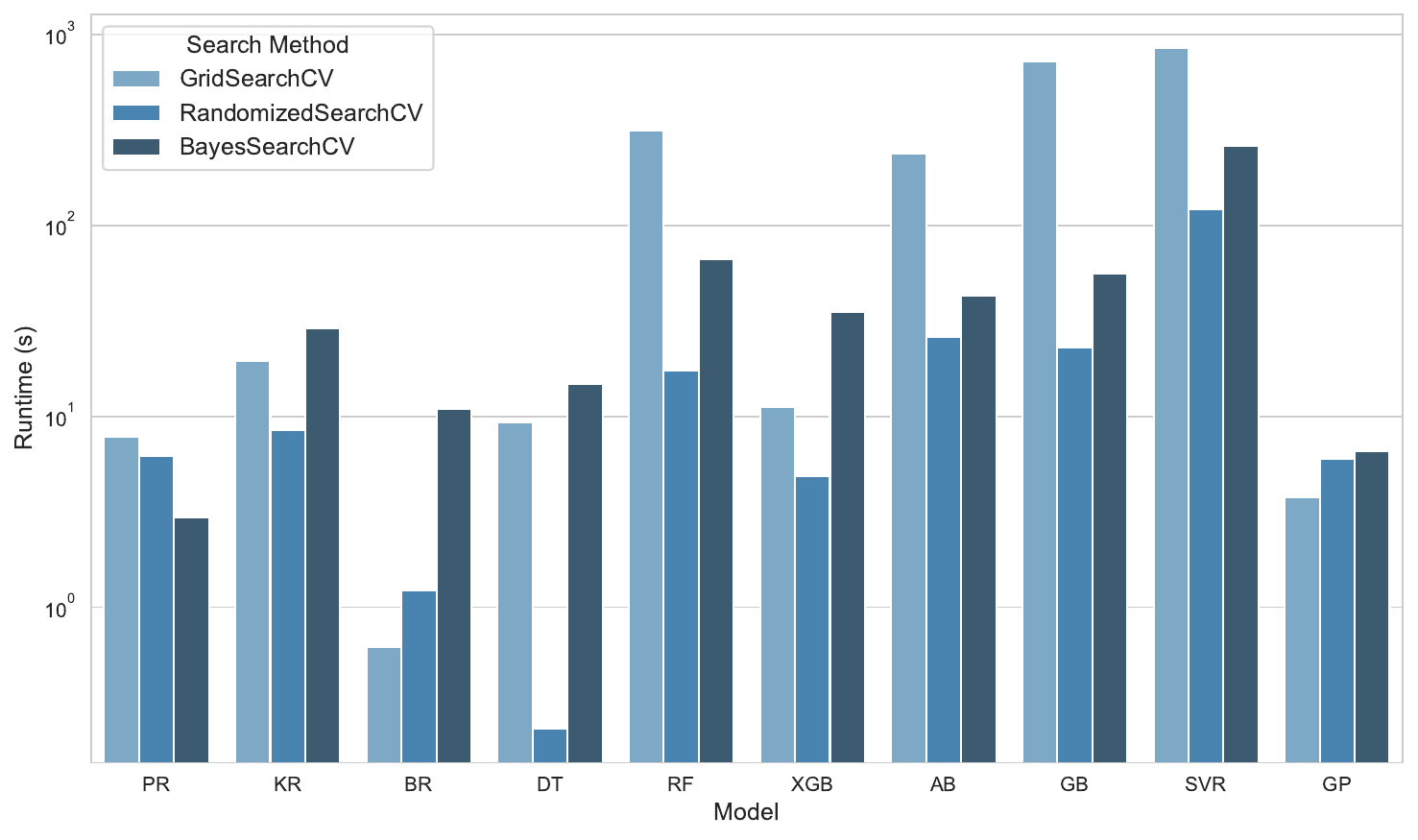}
    \caption{Performance metrics for Aurora}
     \label{fig:aurora}
\end{figure*}
\begin{figure*}[ht]
    \centering
        \includegraphics[width=0.48\textwidth]{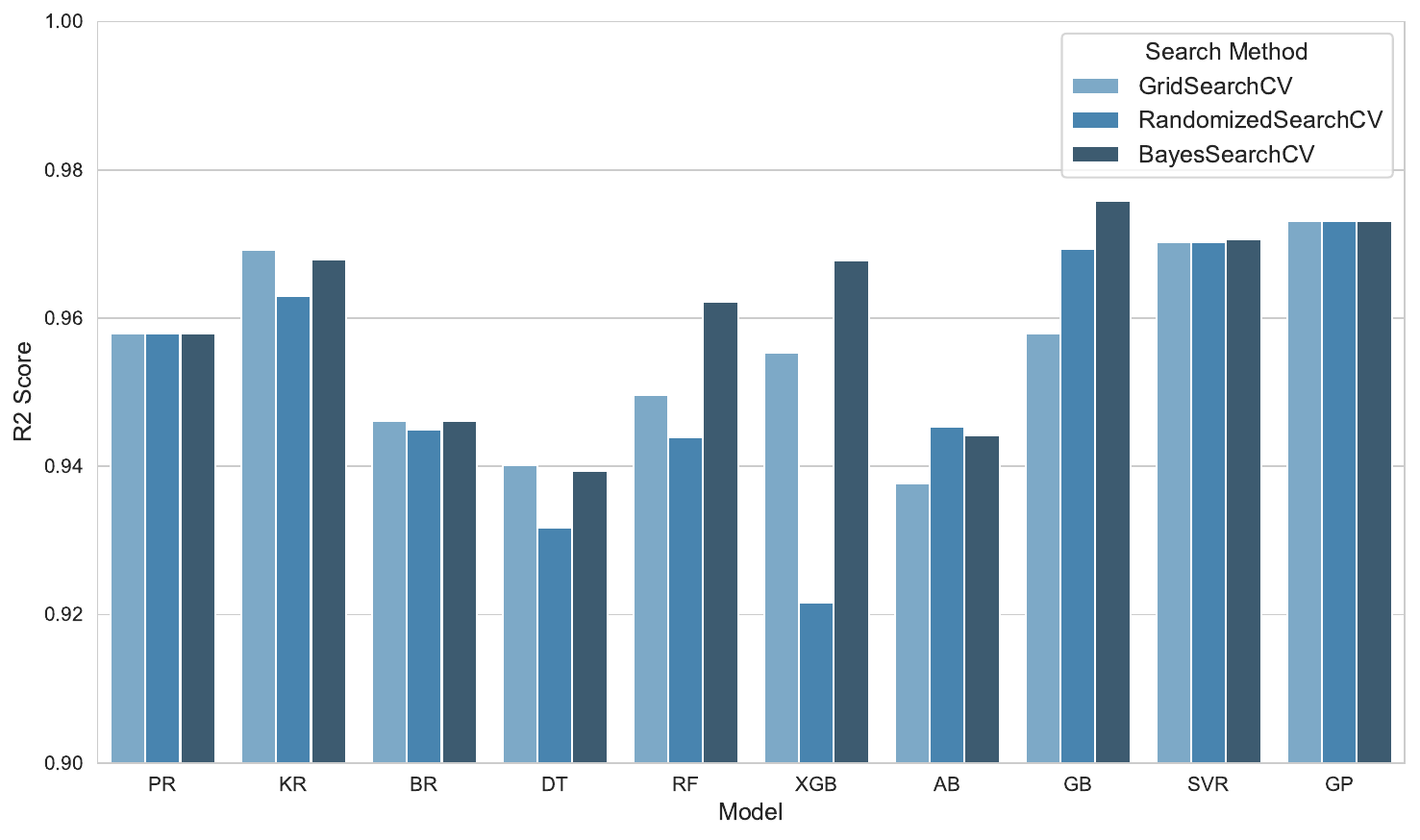}
        \includegraphics[width=0.48\textwidth]{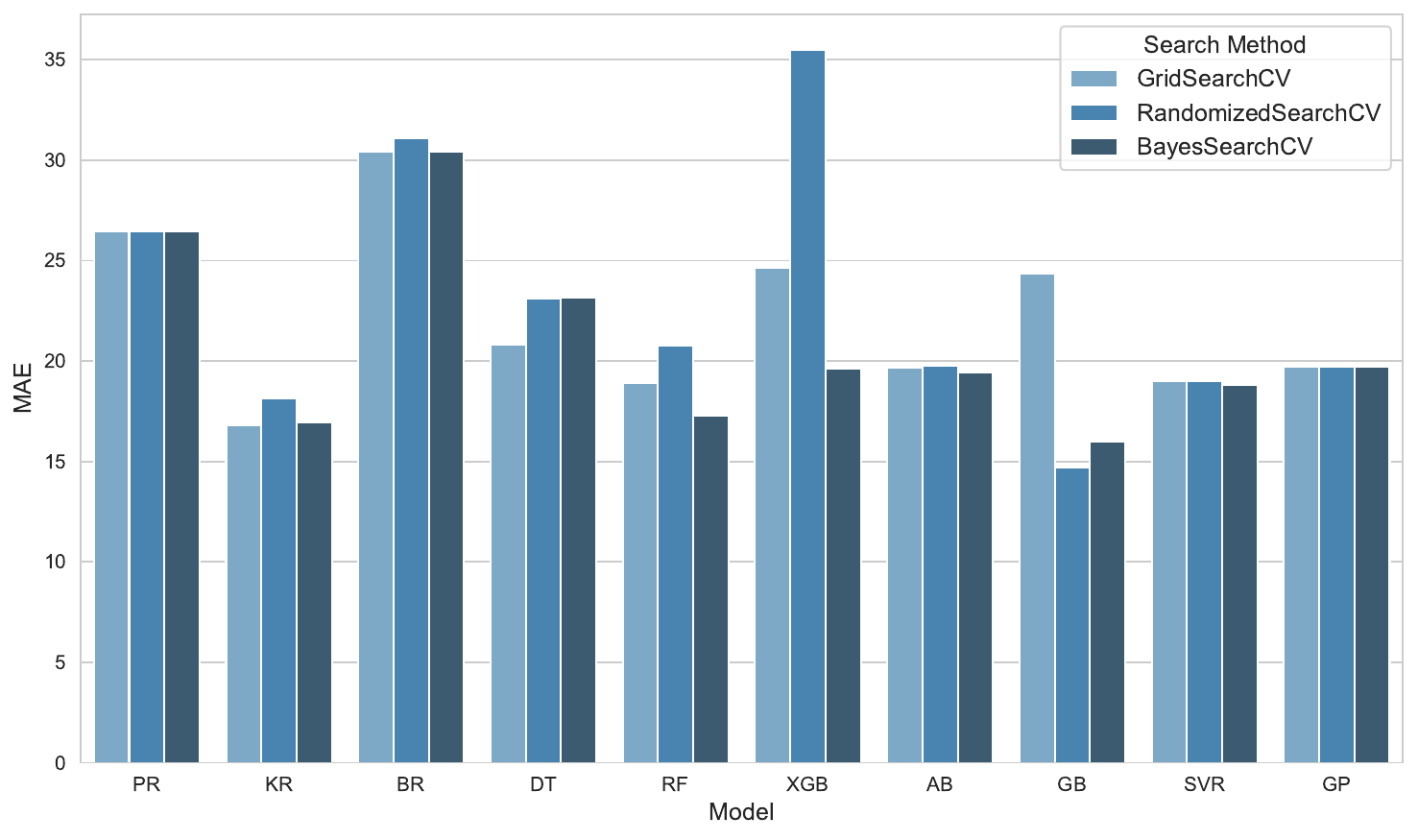}
        \includegraphics[width=0.48\textwidth]{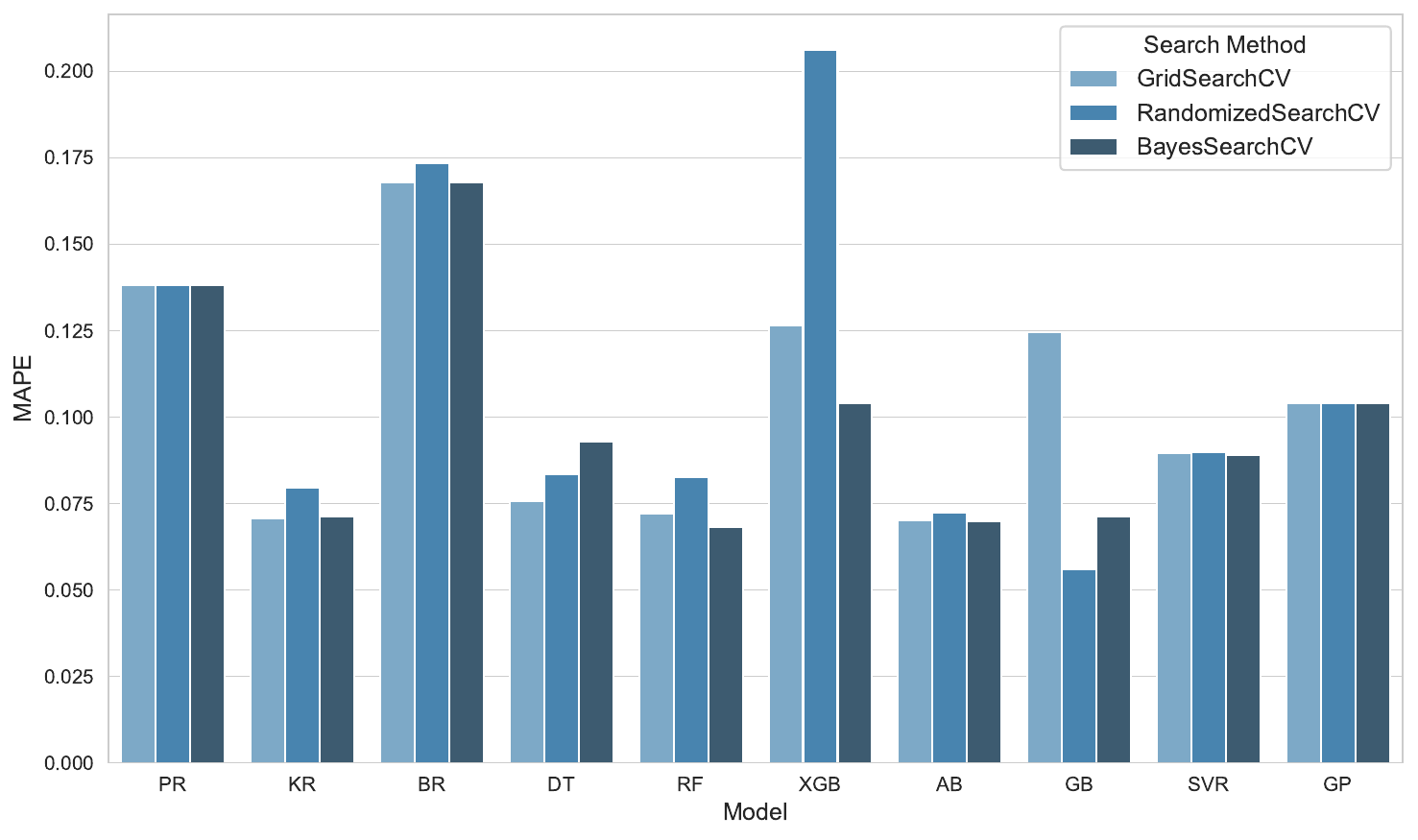}
        \includegraphics[width=0.48\textwidth]{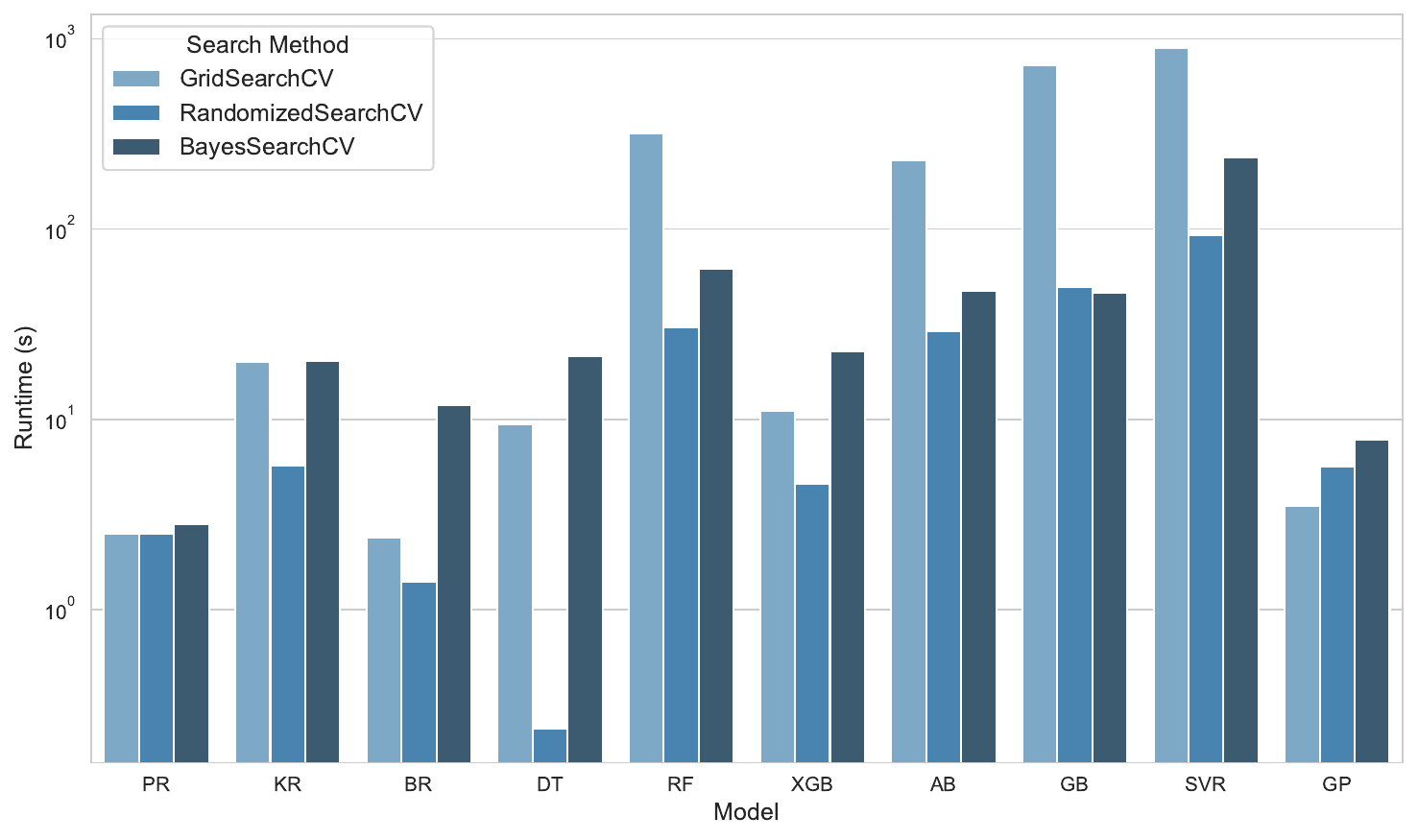}
    \caption{Performance metrics for Frontier}
    \label{fig:frontier}
\end{figure*}

\subsection{Model Results and Initial Analysis}
\label{sub:initial_analysis}

We perform hyper-parameter optimization for the selected
ML models to determine the best performing one. 
Figures \ref{fig:aurora} and \ref{fig:frontier} show the results for $R^2$ score, MAE, MAPE and the optimizations run times for all models and search strategies for Aurora and Frontier respectively. We see that while KR performed the best in the isolated case of MAPE for Aurora optimized with GridSearchCV, GB yields the best overall R2, MAE and MAPE for both machines. As a result, we use GB as our model in active learning with the exception that GP being used for the US strategy.

Table~\ref{tab:trainingtimes} shows the training and prediction times of GB for Aurora and Frontier. 
We see that for both supercomputers, training and prediction takes similar times and on the order of 1.2s and 20ms, respectively. Compared to CCSD execution times they are negligible. As a result of the optimizations, we use GB models with 750 tree-based estimators, a maximum depth of 10, and all other default hyper-parameter values in all the following sections below whenever it is deployed. 

\subsection{Shortest-Time Question (STQ)}
\label{sub:stq}
Table \ref{tab:aurora_shortesttime} shows the parameter configurations
leading to shortest execution times for Aurora. It also shows the predictions
of our ML model for each problem size of $(O, V)$ as described at the end of Section \ref{sub:active-learning}. The model correctly predicts
all but the cases with the predicted execution time shown in parenthesis. 
With the three incorrectly predicted parameter configuration, our model
achieves an $R^2$, MAE, and MAPE score of 0.999, 2.36, and 0.023 respectively.
Even when predicting the incorrect optimal configuration, the configurations our models predicted lead to very similar run times. Similarly, Table \ref{tab:frontier_shortesttime} shows both the parameter configurations
leading to shortest execution times and the predictions
of our ML model for each problem size of $(O, V)$ for Frontier.
We see that in five cases, our model predicts the parameter configuration
incorrectly. However, overall, it achieves an $R^2$, MAE, and MAPE score of 0.969, 4.65, and 0.073 respectively.
\begin{table}
    \centering
    \caption{Training and prediction times for Gradient Boosting.}
    \label{tab:trainingtimes}
    \begin{tabular}{|c|c|c|} \hline
      System   & Training & Prediction \\ \hline
      Aurora   & 1.18 s ± 20.5 ms & 20 ms ± 802 µs  \\ \hline
      Frontier   & 1.19 s ± 1.95 ms & 22.3 ms ± 848 µs  \\ \hline
    \end{tabular}
\end{table}
\begin{table}
    \centering
    \caption{Aurora shortest time results.}
    \label{tab:aurora_shortesttime}
\begin{tabular}{|c|c|c|c|c|}
\toprule
O & V & Nodes & Tile size & Runtime (s) \\
\midrule
44 & 260 & 5 & 40 & {{17.41}} \\
81 & 835 & 185 & 80 & {{66.81}} \\
85 & 698 & 220 & 60 & {{47.05}} \\
99 & 718 & 260 & 60 & {{53.83}} \\
99 & 1021 & 400 & 60 & {{112.70}} \\
116 & 575 & 240(220) & 60 & {{38.35}}({{38.78}}) \\
116 & 840 & 350 & 60 & {{79.95}} \\
116 & 1184 & 400 & 80 & {{180.30}} \\
134 & 523 & 200 & 70 & {{41.86}} \\
134 & 951 & 400 & 70 & {{122.95}} \\
134 & 1200 & 800 & 80 & {{196.70}} \\
146 & 278 & 90 & 70 & {{17.92}} \\
146 & 591 & 120 & 70(80) & {{62.89}}({{66.18}}) \\
146 & 1096 & 300 & 73 & {{186.18}} \\
146 & 1568 & 800(900) & 80 & {{393.72}}({{397.1}}) \\
180 & 720 & 220 & 70 & {{104.36}} \\
180 & 1070 & 320 & 80 & {{232.88}} \\
196 & 764 & 300 & 80 & {{124.95}} \\
204 & 969 & 320 & 90 & {{214.17}} \\
235 & 1007 & 400 & 100 & {{291.99}} \\
280 & 1040 & 110 & 100 & {{605.93}} \\
345 & 791 & 400 & 110 & {{282.83}} \\
\bottomrule
\end{tabular}
\end{table}
\begin{table}
    \centering
    \caption{Frontier shortest time results.}
    \label{tab:frontier_shortesttime}
   \begin{tabular}{|c|c|c|c|c|}
\toprule
O & V & Nodes & Tile size & Runtime (s)\\
\midrule
49 & 663 & 80 & 60 & {{22.24}} \\
81 & 835 & 185 & 80 & {{50.86}} \\
85 & 698 & 75 & 90 & {{56.81}} \\
99 & 718 & 200 & 80 & {{42.24}} \\
99 & 1021 & 200 & 80 & {{108.58}} \\
116 & 575 & 200 & 70 & {{28.54}} \\
116 & 840 & 300(220) & 70(90) & {{59.17(63.07)}} \\
116 & 1184 & 350 & 70 & {{159.66}} \\
134 & 523 & 350(300) & 70(80) & {{25.95(28.9)}} \\
134 & 951 & 400(300) & 70(90) & {{90.15(96.44)}} \\
134 & 1200 & 700 & 80 & {{135.81}} \\
146 & 591 & 70(120) & 90(120) & {{57.82(60.36)}} \\
146 & 1096 & 700 & 90(73) & {{108.84(116.43)}} \\
180 & 720 & 350 & 80 & {{71.43}} \\
180 & 1070 & 400 & 90 & {{172.50}} \\
196 & 764 & 350 & 90 & {{91.41}} \\
204 & 969 & 350 & 100 & {{158.46}} \\
235 & 1007 & 400 & 100 & {{222.96}} \\
280 & 1040 & 600 & 100 & {{249.16}} \\
345 & 791 & 350 & 130 & {{249.66}} \\
\bottomrule
\end{tabular}
\end{table}

\begin{figure*}
    \includegraphics[width=\textwidth]{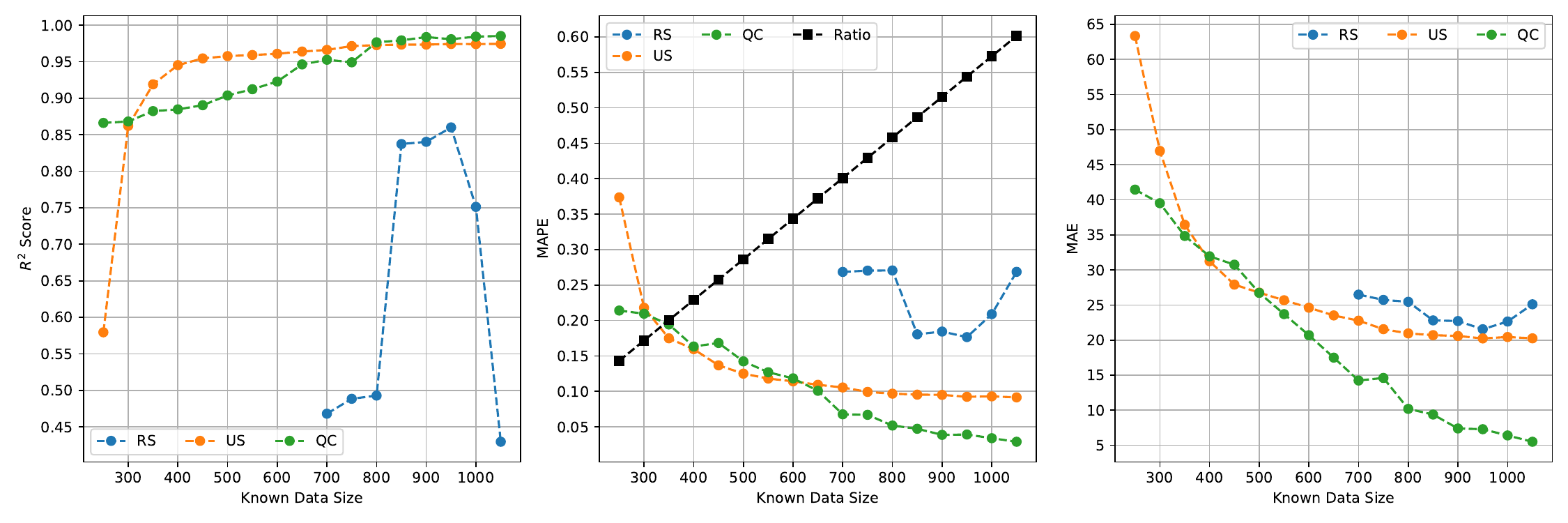}
    \caption{Aurora active learning results.}
    \label{fig:auroraactivelearning}
\end{figure*}
\begin{figure*}
  \centering
  \includegraphics[width=\textwidth]{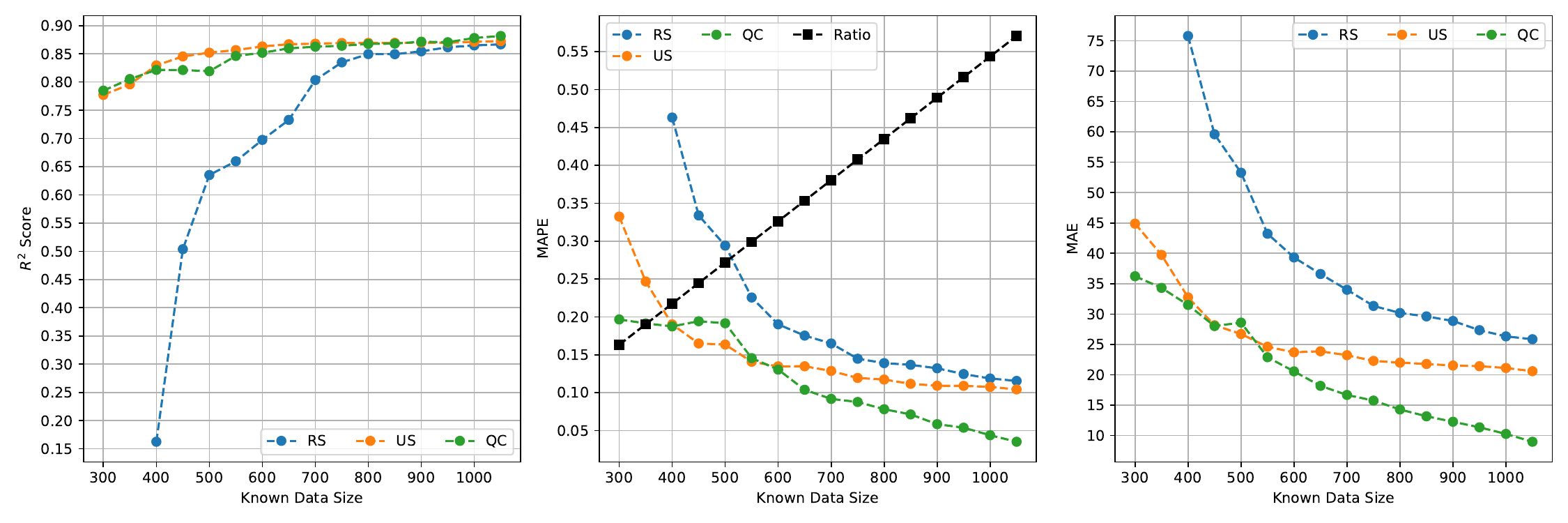}
    \caption{Frontier active learning results.}
    \label{fig:frontieractivelearning}
\end{figure*}

Figures \ref{fig:auroraactivelearning} and 
\ref{fig:frontieractivelearning} show the performances of active learning models as the learning process continue for Aurora and Frontier, respectively. The y-axes shows the $R^2$ score, MAPE, and MAE values while x-axes show the number of known data instances, that is, the results of experiments run on Aurora or Frontier.
There are two types of learning goals with the three types of query strategies. The query strategies are random sampling (RS) as our baseline, uncertainty sampling (US), and query by committee (QC).
The goal of accurately predicting the optimal parameter configuration through accurately predicting execution time is more complex and has inherent performance variance due to imperfect and incomplete information. The goals are the accurate prediction of the execution time and the parameter configurations for the shortest time question (STQ). 

\begin{figure*}
    \includegraphics[width=\textwidth]{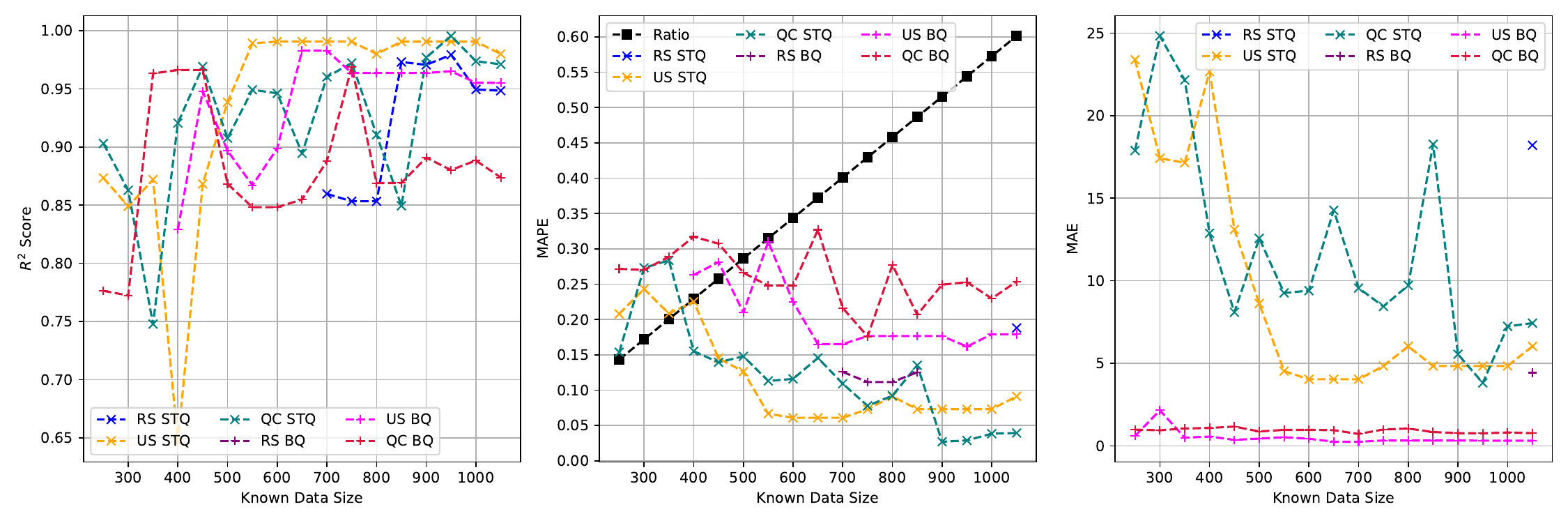}
    \caption{Aurora active learning results for shortest time and budget question.}
    \label{fig:aurorastqbq}
\end{figure*}

In Figure~\ref{fig:aurorastqbq} for Aurora, the baseline RS fails in both goals and we could  plot results only until 700 or more data points are known. $R^2$ scores were negative, and MAPE and MAE were too high to fit to the graph. We do see that the STQ goal introduced performance variance, however, the corresponding models are more successful than the non-STQ models. Moreover,
we see that US performs better than QC overall. \textit{The key observation is that for the STQ goal, a MAPE of about 0.2 is achievable with just around 450 data points, i.e., experiments, (\%25 of the original dataset), a MAPE of 0.1 is achievable with around 550 experiments (\%32 of the original dataset).} The corresponding $R^2$ scores are similarly high and around 0.98. 

\begin{figure*}
  \centering
  \includegraphics[width=\textwidth]{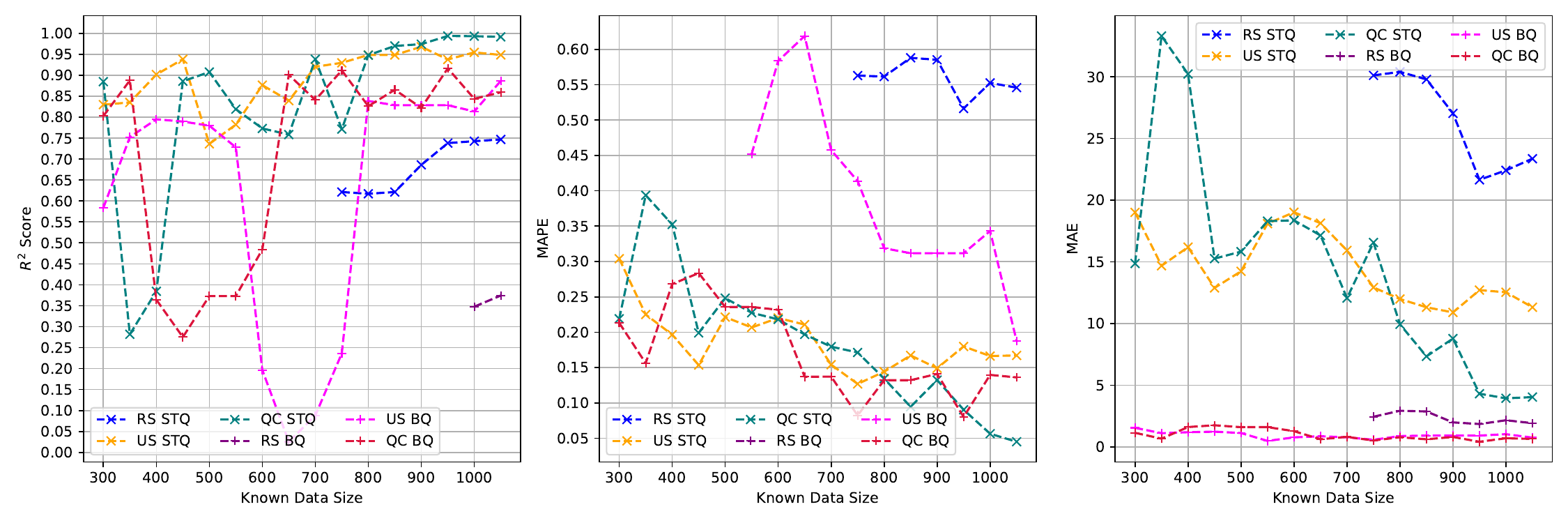}
    \caption{Frontier active learning results for shortest time and budget question.}
    \label{fig:frontierstqbq}
\end{figure*}

In Figure~\ref{fig:frontierstqbq} for Frontier, the baseline RS for the non-STQ goal performs better and more consistently than the models for Aurora. However, the baseline RS-STQ fails just the same as with Aurora. Moreover, the additional performance variance due to the STQ goal is evident as expected. Furthermore, in contrast to Aurora, QC outperforms US in general. As other experiments show, Frontier is more difficult to predict than Aurora. \textit{For Frontier with the STQ goal, our key observation is that a MAPE of about 0.20 is achievable with 450 to 650 experiments, (\%25-35 of the original dataset), and a MAPE of 0.1 is achievable with around 850 experiments (\%47 of the original dataset).}

\begin{table}
    \centering
    \caption{Aurora shortest node hours results.}
    \label{tab:aurora_shortestnodehours}
\begin{tabular}{rrrrrr}
\toprule
Oa & Va & Nodes & tilesize & Runtime(s) & Node Hours \\
\midrule
44 & 260 & 5 & 40 & {{17.41}} & {{0.02}} \\
81 & 835 & 25 & 80 & {{193.26}} & {{1.34}} \\
85 & 698 & 15 & 120 & {{146.45}} & {{0.61}} \\
99 & 718 & 15 & 110(90) & {{173.41}}(182.32) & {{0.72}}(0.76) \\
99 & 1021 & 35 & 110 & {{285.94}} & {{2.78}} \\
116 & 575 & 15 & 90 & {{123.51}} & {{0.51}} \\
116 & 840 & 35 & 90 & {{178.26}} & {{1.73}} \\
116 & 1184 & 15 & 120(140) & {{682.15}}(706.92) & {{2.84}}(2.95)\\
134 & 523 & 65 & 90 & {{58.25}} & {{1.05}} \\
134 & 951 & 35(25) & 130(140) & {{282.70}}(565.37) & {{2.75}}(3.93) \\
134 & 1200 & 45 & 120 & {{469.57}} & {{5.87}} \\
146 & 278 & 10 & 120(100) & {{38.67}}(38.83) & {{0.11}}(0.11) \\
146 & 591 & 30 & 100 & {{102.96}} & {{0.86}} \\
146 & 1096 & 30 & 140 & {{498.74}} & {{4.16}} \\
146 & 1568 & 200 & 90 & {{616.39}} & {{34.24}} \\
180 & 720 & 20 & 130 & {{293.36}} & {{1.63}} \\
180 & 1070 & 30 & 120 & {{591.97}} & {{4.93}} \\
196 & 764 & 50 & 110 & {{247.22}} & {{3.43}} \\
204 & 969 & 90 & 90 & {{380.81}} & {{9.52}} \\
235 & 1007 & 25 & 140 & {{907.16}} & {{6.30}} \\
280 & 1040 & 50(40) & 130(140) & {{876.74}}(1163.77) & {{12.18}}(12.93) \\
345 & 791 & 50 & 130 & {{589.65}} & {{8.19}} \\
\bottomrule
\end{tabular}
\end{table}

\begin{table}
    \centering
    \caption{Frontier shortest node hours results.}
    \label{tab:frontier_shortestnodehours}
\begin{tabular}{rrrrrr}
\toprule
Oa & Va & Nodes & tilesize & Runtime(s) & Node Hours \\
\midrule
49 & 663 & 10 & 150 & {{64.67}} & {{0.18}} \\
81 & 835 & 15 & 150(110) & {{157.31}}(162.3) & {{0.66}}(0.68)\\
85 & 698 & 15 & 110(140) & {{104.37}}(114.63) & {{0.43}}(0.48) \\
99 & 718 & 25 & 130 & {{114.69}} & {{0.80}} \\
99 & 1021 & 35 & 110 & {{206.02}} & {{2.00}} \\
116 & 575 & 15 & 130(180) & {{116.08}}(143.36) & {{0.48}} (0.6)\\
116 & 840 & 35 & 160(150) & {{211.50}}(216.3) & {{2.06}}(2.10) \\
116 & 1184 & 65 & 130 & {{300.80}} & {{5.43}} \\
134 & 523 & 15 & 120 & {{104.93}} & {{0.44}} \\
134 & 951 & 95(35) & 130(160) & {{154.45}}(425.8) & {{4.08}}(4.14) \\
134 & 1200 & 45 & 140 & {{467.01}} & {{5.84}} \\
146 & 591 & 30 & 120 & {{98.69}} & {{0.82}} \\
146 & 1096 & 30(50) & 150(140) & {{532.34}}(426.38) & {{4.44}}(5.92) \\
180 & 720 & 50 & 90 & {{172.48}} & {{2.40}} \\
180 & 1070 & 90 & 150 & {{313.89}} & {{7.85}} \\
196 & 764 & 50 & 150(110) & {{203.36}}(205.37) & {{2.82}}(2.85) \\
204 & 969 & 70(50) & 140(120) & {{323.80}}(463.97) & {{6.30}}(6.44) \\
235 & 1007 & 50 & 150 & {{563.71}} & {{7.83}} \\
280 & 1040 & 70(50) & 140(130) & {{666.34}}(1175.96) & {{12.96}}(16.33) \\
345 & 791 & 50 & 150 & {{606.59}} & {{8.42}} \\
\bottomrule
\end{tabular}
\end{table}


\subsection{Budget Question (BQ)}
Table 5 presents the optimal parameter configurations that yields the smallest node-hours for each problem size $(O, V)$ for Aurora. The model correctly predicts the best configurations except in five cases. The incorrect configurations and their corresponding runtime and node-hours are shown in parenthesis. Despite the five suboptimal configuration predictions, our model achieves $R^2$, MAE, and MAPE scores of 0.979, 0.41 and 0.12.

Similarly for Frontier, Table 6 shows the parameter configurations that lead to the smallest node-hours. There are nine cases where the model predicts suboptimal configurations. Even with these nine incorrect predictions the model achieves $R^2$, MAE, and MAPE scores of 0.892, 0.59, and 0.11.

A key observation from comparing Tables 3 and 5 for Aurora is the model's different behavior in the two different goals. When optimizing for shortest time, the model selects a large number of nodes. When the objective is to minimize users' budget, it consistently selects a smaller number of nodes. Tables 4 and 6 for Frontier show a similar pattern of predicting based on users' priorities. 

For the active learning part on BQ, we again used the same query strategies where we used random sampling (RS) as our baseline, uncertainty sampling (US), and query by committee (QC). Figure~\ref{fig:aurorastqbq} shows the results for Aurora using active learning. \textit{Our key observation for the active learning strategy for the BQ goal is that a MAPE of about 0.2 can be achieved with around 500 experiments using the US strategy, and that a MAPE of close to a 0.15 can be achieved with around 650 experiments. Given that our model achieves a MAPE of 0.12, we could have achieved a close value by using \%35 of the original dataset.} Figure~\ref{fig:frontierstqbq} shows the results for Frontier using active learning. \textit{Our key observation for the Frontier data is that the US strategy is performing better for achieving reasonable MAPE values. Using this strategy we were able to achieve a MAPE score of around 0.15 with about 350 data points, which corresponds to \%20 of the data set we used for our models. } 

Overall, active learning shows great potential for achieving strong predictive performance, as measured by $R^2$, MAE, and MAPE scores, especially in scenarios where data collection is difficult in supercomputing environments. This approach is particularly valuable when training models for new architectures, as it can guide the experimental setup to more efficiently explore the search space for machine learning models.

\section{Conclusions}
\label{conclusions}
We have developed machine learning models for predicting the cost and configuration parameters for running CCSD computations on two modern supercomputers. Our experiments demonstrate that CCSD execution time prediction using a Gradient Boosting model achieves MAPE of 0.023 on Frontier and 0.073 on Aurora. Given the high computational cost of CCSD runs for large molecular systems, such predictive models significantly reduce the need for extensive manual experimentation, thereby lowering both runtime and resource usage to obtain results for a given molecule.
For machines with limited data, we have demonstrated that active learning can be effectively employed to identify the most informative configurations for model training. We have demonstrated that using active learning, our models achieve a MAPE of approximately 0.2 with only around \%25-\%35 of our experimental data set.

\section{Acknowledgments}
This work is supported by the “Transferring exascale computational chemistry to cloud computing environment and emerging hardware technologies (TEC4)” project at the Pacific Northwest National Laboratory (PNNL). The TEC4 project is funded by the Department of Energy (DOE), Office of Science, Office of Basic Energy Sciences program, the Division of Chemical Sciences, Geosciences, and Biosciences (under Grant No. FWP 82037). PNNL is a multi-program national laboratory operated by Battelle Memorial Institute for the U.S. DoE under Contract No. DE-AC06-76RLO-1830.

This research used resources of the Argonne Leadership Computing Facility, a U.S. Department of Energy (DOE) Office of Science user facility at Argonne National Laboratory and is based on research supported by the U.S. DOE Office of Science-Advanced Scientific Computing Research Program, under Contract No. DE-AC02-06CH11357.

This research also used resources of the Oak Ridge Leadership Computing Facility at the Oak Ridge National Laboratory, which is supported by the Office of Science of the U.S. Department of Energy under Contract No. DE-AC05-00OR22725.

\clearpage
\bibliographystyle{plain}
\bibliography{main.bbl}

\end{document}